\documentclass[final]{cvpr}

\usepackage{times}
\usepackage{epsfig}
\usepackage{graphicx}
\usepackage{amsmath}
\usepackage{amssymb}

\usepackage[pagebackref=true,breaklinks=true,colorlinks,bookmarks=false]{hyperref}

\usepackage{multirow}
\usepackage{subfigure}
\usepackage{bm}
\usepackage{multirow}
\usepackage{subfigure}
\usepackage{wrapfig}
\usepackage{hyperref}
\usepackage{url}
\usepackage{cleveref}
\usepackage{gensymb}
\usepackage{booktabs}       

\usepackage{stfloats}

\usepackage{bbm} 

\def\cvprPaperID{4414} 
\def\confYear{CVPR 2021}

\begin{document}
	
	\title{Multi-Task Adversarial Attack}
	
	\author{Pengxin Guo\thanks{The first two authors contributed equally.}, Yuancheng Xu, Baijiong Lin, Yu Zhang
\\
		Department of Computer Science and Engineering, Southern University of Science and Technology\\
		{\tt\small 12032913@mail.sustech.edu.cn, ycxu@umd.edu, linbj@mail.sustech.edu.cn, yu.zhang.ust@gmail.com}
	}
	
	\maketitle
	
	\begin{abstract}
		Deep neural networks have achieved impressive performance in various areas, but they are shown to be vulnerable to adversarial attacks. Previous works on adversarial attacks mainly focused on the single-task setting. However, in real applications, it is often desirable to attack several models for different tasks simultaneously. To this end, we propose \textbf{M}ulti-\textbf{T}ask adversarial \textbf{A}ttack (MTA), a unified framework that can craft adversarial examples for multiple tasks efficiently by leveraging shared knowledge among tasks, which helps enable large-scale applications of adversarial attacks on real-world systems. More specifically, MTA uses a generator for adversarial perturbations which consists of a shared encoder for all tasks and multiple task-specific decoders. Thanks to the shared encoder, MTA reduces the storage cost and speeds up the inference when attacking multiple tasks simultaneously. Moreover, the proposed framework can be used to generate per-instance and universal perturbations for targeted and non-targeted attacks. Experimental results on the Office-31 and NYUv2 datasets demonstrate that MTA can improve the quality of attacks when compared with its single-task counterpart.


		
	\end{abstract}
	
	\section{Introduction}
	
	\begin{figure*} [!htb]
		\centering
		\subfigure[Optimization-based and iterative methods]{\includegraphics[width=0.33\textwidth]{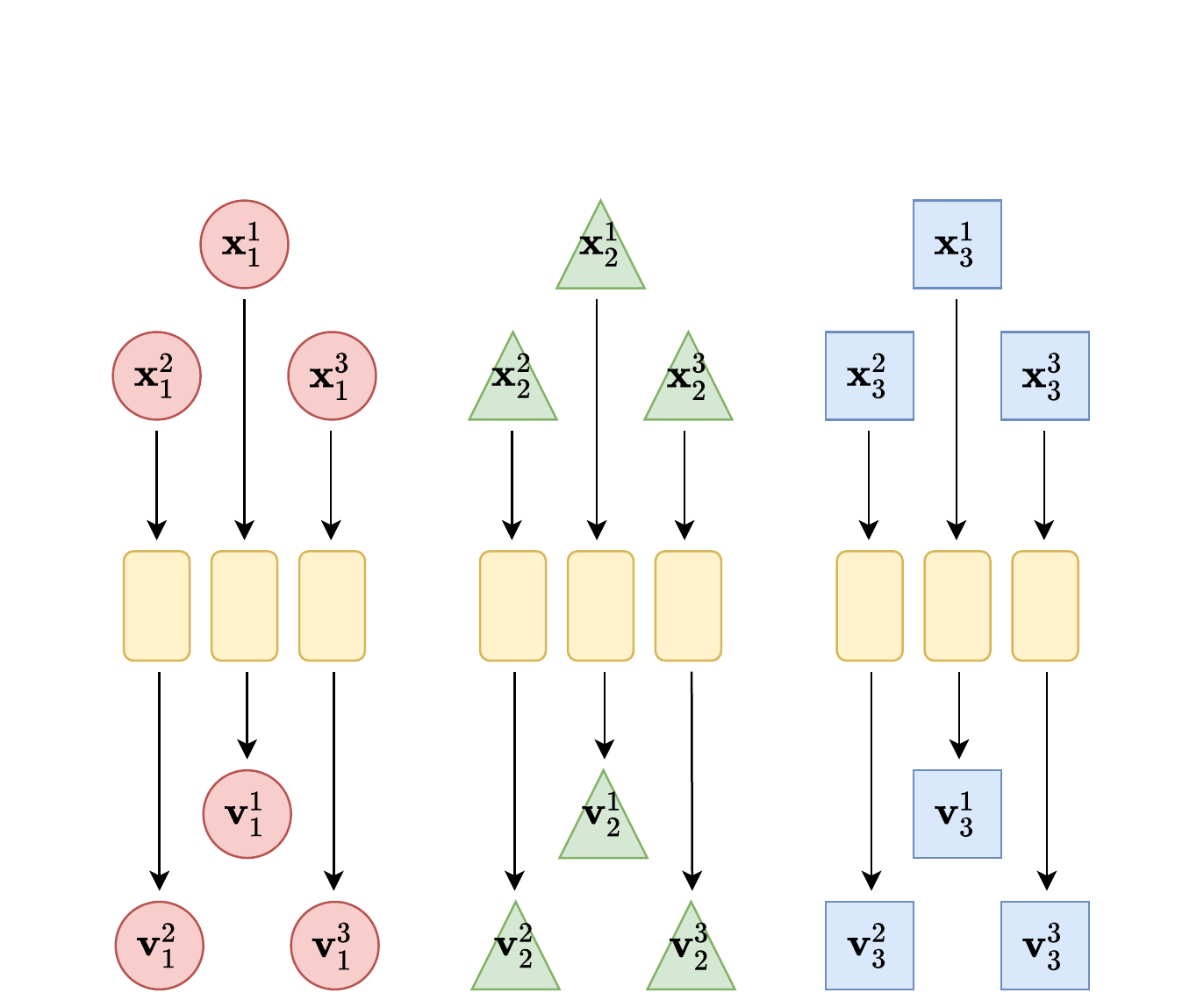}{\label{fig: motivation_opt}}}
		\subfigure[Single-task generative methods]{\includegraphics[width=0.33\textwidth]{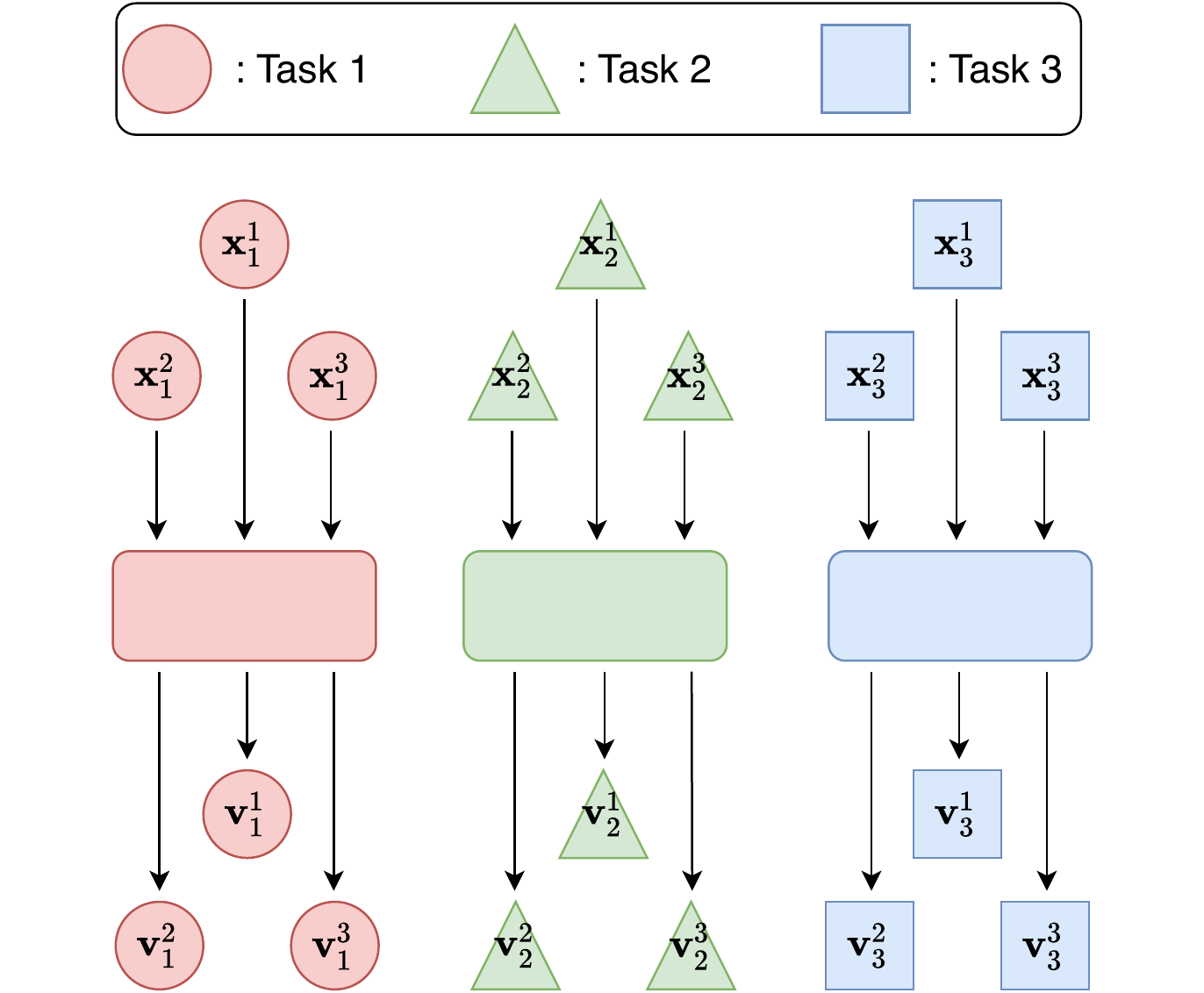}{\label{fig: motivation_generative}}}
		\subfigure[Multi-task adversarial attack (MTA)]{\includegraphics[width=0.33\textwidth]{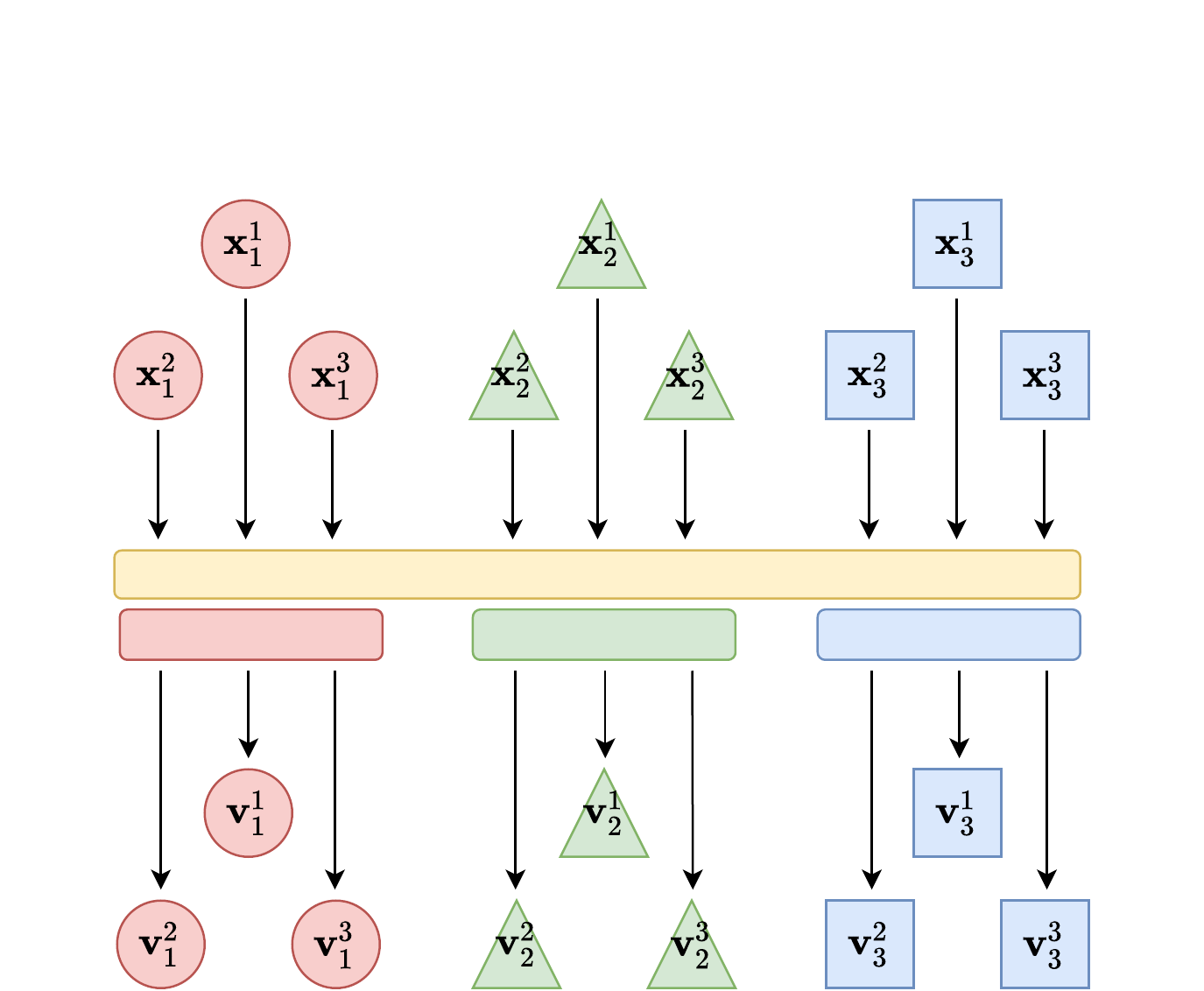}{\label{fig: motivation_MTA}}}
		\caption{Three different schemes for crafting adversarial examples. $x_{t}^{i}$ is the $i$-th instance of task $t$ and $v_{t}^{i}$ is the corresponding adversarial example. (a) Optimization-based and iterative methods craft adversarial examples for each instance one at a time. (b) Single-task generative methods learn to generate adversarial examples from the whole training data, but they do not consider relatedness among tasks. (c) The proposed MTA framework exploits shared knowledge among tasks to craft adversarial examples for all tasks.}
		\label{fig: motivation}
	\end{figure*}
	
	
	Although deep learning has achieved impressive performance on a range of computer vision tasks, it is vulnerable to adversarial examples \cite{DBLP:conf/pkdd/BiggioCMNSLGR13,DBLP:journals/corr/SzegedyZSBEGF13, DBLP:journals/corr/GoodfellowSS14}, which are crafted by adding human-imperceptible perturbations to clean data in order to mislead neural network models.

	Previous works on adversarial attacks have focused on the single-task setting where adversarial examples are crafted on one dataset for a single task. However, in real-world applications, an attacker may wish to craft adversarial examples for several related tasks simultaneously in order to harm the performance of multiple systems more efficiently. For example, since modern computer vision systems rely heavily on deep neural networks, it is often desirable to efficiently attack multiple building blocks of the systems, such as models for image classification, semantic segmentation, depth estimation and so on. Therefore, it is important to consider adversarial attacks in the multi-task setting,
	which helps enable large-scale applications of adversarial attacks on real world systems.
	
	However, existing attack methods are not optimized for the multi-task setting. For example, current iterative and optimization-based attack methods \cite{DBLP:journals/corr/GoodfellowSS14,DBLP:conf/cvpr/Moosavi-Dezfooli16,DBLP:conf/iclr/MadryMSTV18} craft adversarial examples for each instance one at a time by solving an optimization problem particular to that instance. Therefore, the process of crafting adversarial examples for different instances is independent of each other (Figure \ref{fig: motivation_opt}). On the other hand, current generative attack methods \cite{DBLP:conf/ijcai/XiaoLZHLS18,
		DBLP:journals/corr/BalujaF17,DBLP:conf/cvpr/PoursaeedKGB18} train neural networks (\ie generators) to generate adversarial examples for all the instances of the dataset for a single task.
	However, these methods train generators independently on each task (Figure \ref{fig: motivation_generative}) without exploiting the shared knowledge among tasks that might be useful for more efficient generation of adversarial examples of higher quality.
	
	To solve the aforementioned problems, in this paper we extend adversarial attacks to the multi-task setting and propose \textbf{M}ulti-\textbf{T}ask adversarial \textbf{A}ttack (MTA), a unified framework that can craft adversarial examples for multiple tasks efficiently by leveraging the shared knowledge among tasks. The scheme of MTA is illustrated in Figure \ref{fig: motivation_MTA}. MTA employs an end-to-end trainable generator with parameter sharing that learns to generate adversarial examples by exploiting the relatedness among tasks. Compared with its single-task counterpart, MTA improves the quality of adversarial examples and also reduces the storage cost as well as the inference time, enabling large-scale generations of adversarial examples for multiple tasks.
	
	Moreover, MTA is flexible as it can generate per-instance and universal adversarial perturbations for targeted and non-targeted attacks. To explain these types of attacks, there are two ways to categorize adversarial perturbations. Firstly, they can be categorized into universal perturbations, which can be added to any input, and per-instance perturbations, which depend on the inputs. Secondly, adversarial attacks can be targeted and non-targeted. The goal of non-targeted attacks is to decrease the overall performance of the pre-trained model, while targeted attacks aim to change the predictions of the pre-trained model on adversarial examples to some classes specified by the attacker. Therefore, along these two categorizations, MTA is able to generate four possible types of adversarial perturbations as mentioned above.

	The main contributions of this paper are summarized as follows.
	\begin{itemize}
		\item To the best of our knowledge, we are the first to extend adversarial attacks to the multi-task setting by learning an end-to-end trainable generator with parameter sharing among tasks. The proposed MTA model can generate high-quality adversarial examples for multiple tasks simultaneously.
		\item The proposed MTA framework is flexible as it can efficiently generate per-instance and universal perturbations for both targeted and non-targeted attacks.
		\item Experimental results show that MTA improves the quality of attacks, reduces the storage cost, and achieves faster inference when compared with single-task generative approaches to adversarial attacks.
		
	\end{itemize}

	\section{Related Work}
	
	\subsection{Adversarial Attack}
	\label{sec:related_work_attack}
	Adversarial attacks aim to fool pre-trained models by manipulating the input data, such as adding imperceptible noises. In this paper, we consider evasion attacks where an input is perturbed at the inference phase. In this section, we review several representative attack methods.
	
	Optimization-based non-targeted per-instance attack methods craft adversarial perturbations by solving
	\begin{equation}\label{eq: opt adv attack}
	\max_{v}\ \   \ell(x+v,y,\theta)\quad \text{s.t.} \quad ||v||_{p} \leq \epsilon
	\end{equation}
	where $x$ is the clean input data (\eg a natural image), $y$ is the ground truth label of $x$, $v$ is the adversarial perturbation to be learned, $\theta$ is the set of parameters of the pre-trained model, $\ell$ is a proxy loss of interest (\eg cross-entropy for classification problems), $\| \cdot \|_p$ denotes the $L_p$ norm, and $\epsilon$ is the perturbation threshold. Problem (\ref{eq: opt adv attack}) corresponds to the per-instance attacks since the perturbation $v$ is dependent on the input $x$. Moreover, problem (\ref{eq: opt adv attack}) is to find a bounded perturbation in order to maximize the loss between the prediction on the perturbed input and the ground truth, and therefore it belongs to the non-targeted attacks. Many methods have been proposed to find approximate solutions to
	problem (\ref{eq: opt adv attack}) or its variants. For example, the Fast Gradient Sign Method (FGSM) \cite{DBLP:journals/corr/GoodfellowSS14} proposes to use  $\epsilon\cdot\text{sign}(\nabla_{x}\ell(x,y,\theta))$ as the perturbation $v$, where $\nabla_{x}$ denotes the operator to compute the gradient with respect to $x$ and $\text{sign}(\cdot)$ denotes the elementwise sign function. Other popular methods include DeepFool  \cite{DBLP:conf/cvpr/Moosavi-Dezfooli16} and Projected Gradient Descent (PGD) \cite{DBLP:conf/iclr/MadryMSTV18}. Note that in problem (\ref{eq: opt adv attack}), if $y$ is replaced by an output specified by the attacker and we change to minimize $\ell(x+v,y,\theta)$ with respect to $v$, it corresponds to the targeted attacks.
	
	Unlike per-instance perturbations, universal perturbations, first introduced in \cite{DBLP:conf/cvpr/Moosavi-Dezfooli17}, can be directly added to any test instance to fool the pre-trained model. Moosavi-Dezfooli \etal  \cite{DBLP:conf/cvpr/Moosavi-Dezfooli17} propose an iterative algorithm to find a bounded universal perturbation by iterating over samples to compute a minimal adversarial perturbation for each instance, followed by aggregating per-instance perturbations and projecting the result onto the $\epsilon$-ball around the origin.
	
	Another streamline of works focuses on constructing adversarial examples using generative models. In per-instance attacks, generative methods construct adversarial examples via a generative model, which is more efficient than optimization-based and iterative methods at the inference phase, since the latter needs to solve optimization problems for each test instance. Several generative approaches to adversarial attacks have been proposed. For example, Xiao \etal \cite{DBLP:conf/ijcai/XiaoLZHLS18} propose to use GAN to produce adversarial examples with a high perceptual quality. Baluja \etal \cite{DBLP:journals/corr/BalujaF17} propose to train a generator for adversarial examples by using a loss function that promotes high similarities in the input space and high dissimilarities in the output space. The most relevant work to ours is the Generative Adversarial Perturbations (GAP) method \cite{DBLP:conf/cvpr/PoursaeedKGB18}, which presents a generative neural network that can produce per-instance and universal perturbations for targeted and non-targeted attacks. However, \cite{DBLP:conf/cvpr/PoursaeedKGB18} only considers single-task adversarial attacks, while we focus on multi-task adversarial attacks. The proposed MTA method can efficiently generate adversarial examples for multiple tasks at the same time.
	
	\subsection{Multi-Task Learning}
	Multi-task learning \cite{caruana97,zy17b} leverages shared knowledge contained in multiple related tasks to improve their performance. Zhang and Yang \cite{zy17b} point out three issues to be addressed in multi-task learning: when to share, what to share, and how to share. The ``when to share" problem requires to make decisions between single-task and multi-task learning models for a multi-task learning problem. Since multi-task learning models may suffer from the `negative transfer' phenomenon, deciding whether or not to use multi-task models is important  \cite{DBLP:conf/icml/LeeYH18,DBLP:conf/aaai/CaoPZYY10}. The ``what to share" issue is about determining the form of the shared knowledge such as features or parameters. Lastly, ``how to share" specifies concrete ways to share knowledge. For example, the low-rank approaches  \cite{DBLP:journals/jmlr/AndoZ05,DBLP:conf/icml/ChenTLY09,DBLP:conf/nips/AgarwalDG10} assume that the relatedness among tasks implies the low-rank structure in parameters and thus penalizing large rank of parameter matrices in the objective functions. In task relation learning approaches  \cite{DBLP:conf/uai/ZhangY10,DBLP:conf/nips/BonillaCW07,DBLP:conf/icml/LeeYH18}, the task relatedness is quantified by the similarity or correlation that is learned automatically from data. Task clustering approaches \cite{DBLP:journals/corr/abs-1905-07553,DBLP:conf/icml/ThrunO96} extend clustering methods to the task level and use the same models for tasks within a cluster.
	
	In this paper, we focus on one of the most commonly used approaches in deep multi-task learning \cite{caruana97, DBLP:conf/cvpr/Kokkinos17,DBLP:conf/cvpr/LuKZCJF17}, where different tasks share the first several hidden layers as the shared encoder and then have task-specific parameters in the following layers as decoders. Since the encoder is trained on several tasks, it has the potential to generalize better on multiple tasks. Other potential advantages of solving several tasks jointly instead of independently include lower inference and training time, reducing storage cost, and increased data efficiency. Inspired by these benefits of multi-task learning, we propose to extend adversarial attacks to the multi-task learning setting by using multi-task learning techniques to train a generator.
	
	There has been few work on adversarial attacks under the multi-task learning setting. A recent work \cite{mgnrsyv20} points out that the adversarial robustness of deep neural networks increases as the number of tasks increases. This work differs from ours in that it considers to defense optimization-based and iterative attack methods such as FGSM and PGD which are for non-targeted attacks with the universal perturbation. On the contrary, the proposed MTA focuses on attacks and presents generative models for universal and per-instance perturbations for targeted and non-targeted attacks that can learn different perturbations for each task simultaneously.
	
	\begin{figure*}[!htbp]
		\centering
		\includegraphics[width=0.8\textwidth]{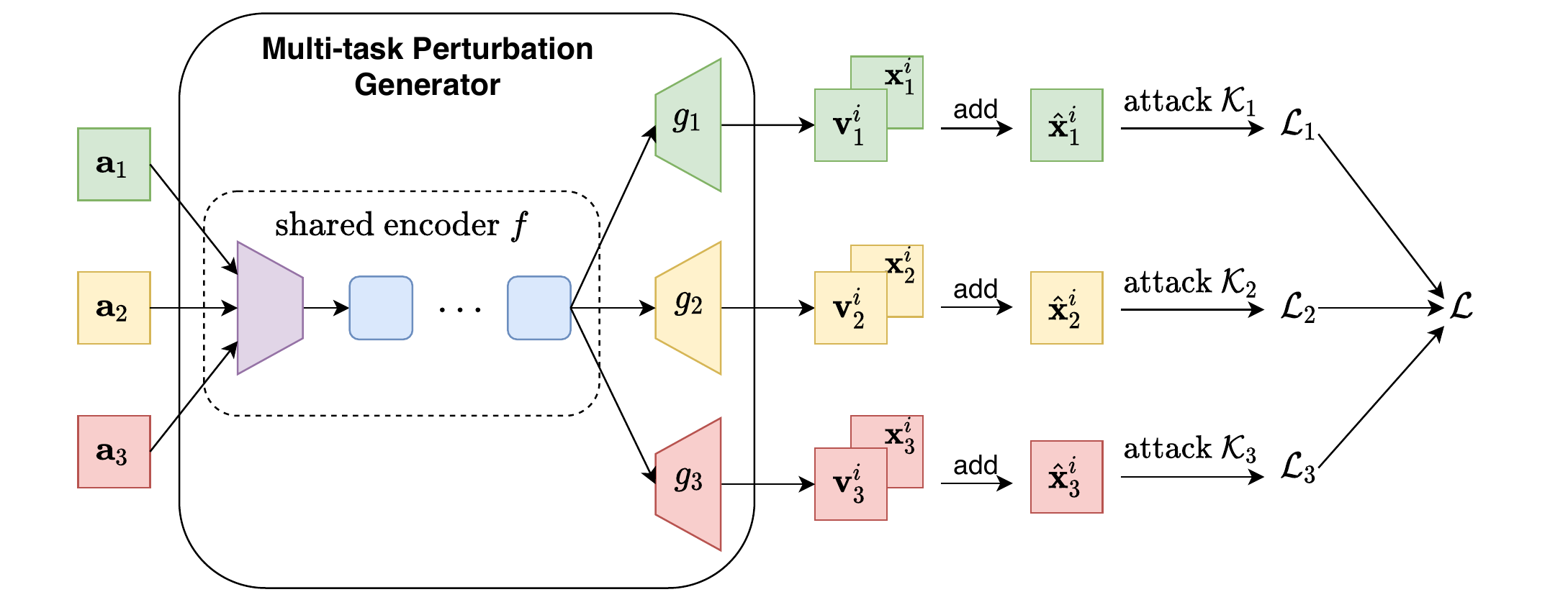}
		\caption{Pipeline for training the generator for multi-task perturbations. The number of tasks $M$ is set to three for illustration. When generating universal perturbations, the input of the generator $a_{t}$ is a random pattern $\mathcal{Z}_{t}$ for each task $t$. When generating per-instance perturbations, $a_{t} = x_{t}^i$ which is an input image. The input $a_{t}$ is first processed by the shared encoder $f$ consisting of downsampling layers followed by several residual blocks. Then task-specific decoder $g_{t}$ is applied to $f(a_{t})$ and after scaling the result to have $L_p$ norm no greater than the perturbation threshold $\epsilon$, we obtain perturbation $v_{t}^i$.
			In particular, for the universal perturbations, $v_t^i = v_t^j = v_t$, because they share the same input $\mathcal{Z}_t$. Then the adversarial example $\hat{x}_{t}^i = x_t^i + v_t^i$ is fed to the pre-trained model $\mathcal{K}_{t}$ and the fooling loss $\mathcal{L}_{t}$ is calculated. The total objective $\mathcal{L}$ for training the generator is a weighted sum of fooling losses $\mathcal{L}_{t}$. Depending on the objective, $\mathcal{L}$ can be targeted or non-targeted.}
		\label{fig:architechture}
	\end{figure*}
	
	\section{Multi-Task Adversarial Attack}
	
	In this section, we introduce the proposed MTA, a framework to generate adversarial examples for all tasks by learning a multi-head generator jointly on all the datasets and pre-trained models associated with multiple tasks. We will show that MTA is flexible in that it can generate universal and per-instance perturbations for targeted and non-targeted attacks.
	
	Suppose that there are $M$ models $\{\mathcal{K}_{t}\}_{t=1}^{M}$ for $M$ tasks, where $\mathcal{K}_{t}$ is trained on a clean dataset $\mathcal{D}_t = \{ (x_t^i, y_t^i)\}_{i=1}^{n_t}$, $n_t$ denotes the number of training data in task $t$, and $y_t^i \in \{1,\cdots,C_{t}\}$ corresponds to a $C_t$-class classification problem for task $t$.
	Given $\mathcal{K}_t$, we say that $\hat{x}_t^i$ is an $\epsilon$-adversarial example of $x_t^i$ if $\mathcal{K}_t(\hat{x}_t^i) \neq y_{t}^i$ and $d(x_t^i,\hat{x}_t^i)\leq \epsilon$, where $d$ is a distance metric and $\epsilon$ is the perturbation threshold.
	
	Note that our formulation includes several special settings. Firstly, $\{\mathcal{K}_{t}\}_{t=1}^{M}$ can be either trained independently or trained jointly with multi-task learning models. Secondly, datasets $\{D_{t}\}_{t=1}^{M}$ can have different inputs but share the same label space such as in the Office-31 dataset \cite{saenko2010adapting}, 
	or share the same inputs but have different label spaces
	such as in the NYUv2 dataset \cite{silberman2012indoor}. 
	Moreover, our framework is not limited to classification tasks. For example, in a semantic segmentation task where each pixel in an $h\times w$ image needs to be classified, the predicted label is $\mathcal{K}_t(x_t^i) = (\mathcal{K}_t(x_{t}^{i})^1,\ldots,\mathcal{K}_t(x_{t}^{i})^n)\in \{1,\ldots,C_t\}^{n}$ for an image $x_t^i$ with the ground truth label $y_t^i = (y_t^{i,1},\ldots,y_t^{i,n})$, where $n = h\cdot w$. Also, MTA can deal with regression problems, such as depth estimation and surface normal estimation (Section \ref{sec: NYU results}). In this section, we focus on classification problems for illustration.
	
	
	
	
	\subsection{Universal Perturbations}
	
	According to \cite{DBLP:conf/cvpr/Moosavi-Dezfooli16}, $v_t$ is a non-targeted universal perturbation for task $t$ if for most $x_t^i \sim \mu_t$ where $\mu_t$ is the data distribution of task $t$, $\mathcal{K}_t(x_t^i + v_t) \neq \mathcal{K}_t(x_t^i)$ holds. Here $v_t$ can be directly added to any test image to fool the model $\mathcal{K}_t$. $v_t$ is required to satisfy $\|v_t\|_p \leq \epsilon$ as it is an imperceptible perturbation. Different from the iterative approach to construct universal perturbations as in \cite{DBLP:conf/cvpr/Moosavi-Dezfooli16}, the proposed MTA is to learn $\{v_t\}_{t=1}^M$ simultaneously in an end-to-end fashion by exploiting the shared knowledge among tasks. To this end, we seek to find a shared encoder $f$ and task-specific decoders $\{g_t\}_{t=1}^M$ such that $g_t(f(\cdot))$ can map a random pattern $\mathcal{Z}_t$ to a universal perturbation $v_t$ for task $t$.
	
	The architecture of the proposed MTA framework  is illustrated in Figure \ref{fig:architechture}. The generator in MTA mainly consists of two parts: the shared encoder $f$ and task-specific decoders $\{g_t\}_{t=1}^{M}$. Similar to \cite{DBLP:conf/cvpr/PoursaeedKGB18}, we adopt the ResNet generator \cite{DBLP:conf/eccv/JohnsonAF16,DBLP:conf/iccv/ZhuPIE17} which consists of several downsampling layers, residual blocks, and upsampling layers. To adapt the ResNet generator to the multi-task setting, the downsampling layers and residual blocks serve as shared encoder $f$ and upsampling layers are used as task-specific decoders $\{g_t\}_{t=1}^{M}$. A universal perturbation $v_{t}$ for task $t$ is obtained by first sampling a random pattern $\mathcal{Z}_{t}$ (denoted by $a_{t}$ in Figure \ref{fig:architechture}) from a uniform distribution, then feeding it to the generator $g_t(f(\cdot))$, and finally normalizing it to have the $L_{p}$ norm no greater than the perturbation threshold $\epsilon$. That is, $v_{t} = \Pi_{\epsilon}(g_t(f(\mathcal{Z}_{t})))$ where $\Pi_{\epsilon}$ is the scaling map which multiplies $g_t(f(\mathcal{Z}_{t}))$ by $\min (1, \frac{\epsilon}{||g_t(f(\mathcal{Z}_{t}))||_{p}})$ . $\Pi_{\epsilon}$ will be omitted for simplicity.
	
	The perturbed input of $x_{t}^i$ is denoted by $\hat{x}_{t}^i = x_{t}^i + v_{t}$, where $v_{t} = g_t(f(\mathcal{Z}_{t}))$. Let $k_t(x_t^i)$ denote the output probabilities of the model $\mathcal{K}_t$ and let $\mathbbm{1}_{y_t^i}$ denote the one-hot encoding of the ground truth label $y_t^i$ for the input $x_t^i$. The goal of non-targeted attacks is to make the prediction on adversarial examples different from the ground truth label. That is, given $\mathcal{K}_t$, we would like the cross-entropy loss $\mathcal{H}(k_t(\hat{x}_t^i), \mathbbm{1}_{y_t^i})$ to be as large as possible. Therefore, we use the following fooling loss for the non-targeted universal perturbations of task $t$ as
	
	\begin{equation} \label{eq: non-targeted loss}
	\ell^{\mathrm{non-tar}}_{t} = \frac{1}{n_t}\sum_{i=1}^{n_t}-\log(\mathcal{H}(k_{t}(\hat{x}_{t}^i), \mathbbm{1}_{y_t^i})).
	\end{equation}
	Then the generative model $(f,\{g_{t}\}_{t =1}^{M})$ for non-targeted universal perturbations is trained jointly by minimizing the weighted sum of all task losses as
	\begin{equation}\label{eq: total non-targeted}
	\mathcal{L}_{\mathrm{non-tar}} = \sum_{t=1}^M \alpha_{t}\ell^{\mathrm{non-tar}}_{t},
	\end{equation}
	where $\alpha_t$ is the weight of task $t$ and satisfies $\sum_{t=1}^M \alpha_{t} = 1$. Without a prior knowledge, $\alpha_t$ can be simply set to $\frac{1}{M}$. We find that the $\log(\cdot)$ function in Eq. \eqref{eq: non-targeted loss} is important to the attack performance in experiments, since it alleviates the domination among tasks by reducing difference in scaling among $\{\ell^{\mathrm{non-tar}}_{t}\}_{t=1}^M$.
	
	For targeted universal perturbations, given a target class $T_{t}$ for each task $t$, its goal is to fool $\mathcal{K}_{t}$ into classifying all inputs from task $t$ as the target class $T_{t}$. Therefore, we formulate the fooling loss for task $t$ as
	\begin{equation}\label{eq: targeted loss}
	\ell^{\mathrm{tar}}_{t} = \frac{1}{n_t}\sum_{i=1}^{n_t}\log(\mathcal{H}(k_{t}(\hat{x}_{t}^i), \mathbbm{1}_{T_{t}})).
	\end{equation}
	Then the objective function of the generative model $(f,\{g_{t}\}_{t =1}^{M})$ for the universal targeted attacks is formulated as
	\begin{equation} \label{eq: total targeted}
	\mathcal{L}_{\mathrm{tar}}= \sum_{t=1}^M \alpha_{t}\ell^{\mathrm{tar}}_{t},
	\end{equation}
	where $\alpha_t$ is the weight of task $t$ and satisfies $\sum_{t=1}^M \alpha_{t} = 1$.
	
	\subsection{Per-instance Perturbations} \label{sec: per-instance pert}
	
	Unlike universal perturbations, per-instance perturbations depend on input instances. As shown in Figure \ref{fig: motivation}, previous optimization-based per-instance attack methods such as FGSM and PGD find perturbations for each instance independently by solving an optimization problem particular to that instance. Previous generative approaches for adversarial attacks train generators independently on each task. These methods are limited to the single-task setting. Unlike these approaches, we would like to jointly train a generator for perturbations for all tasks by parameter sharing. The generator should map an input instance to its additive imperceptible perturbation.
	
	To achieve this, we seek to learn a shared encoder $f$ and task-specific decoders $\{g_{t}\}_{t=1}^{M}$ such that $g_t(f(\cdot))$ maps an input image $x_{t}^i$ to its perturbation $v_{t}^i$. The architecture is shown in Figure \ref{fig:architechture}. The input instance $x_{t}^i$ (denoted by $a_{t}$ in Figure \ref{fig:architechture}) of task $t$ is fed to the shared encoder $f$ followed by the task-specific decoder $g_{t}$ to create the perturbation. Then the perturbation is scaled to have the $L_p$ norm no greater than the perturbation threshold $\epsilon$. The perturbed input $\hat{x}_{t}^i$ is computed as $\hat{x}_{t}^i = x_{t}^i + v_{t}^i$, where $v_{t}^i = g_t(f(x_{t}^i))$.\footnote{Note that we omit the scaling map $\Pi_{\epsilon}$ here. Rigorously, $v_{t}^i = \Pi_{\epsilon}(g_t(f(x_{t}^i))) = g_t(f(x_{t}^i)) \cdot \min (1, \frac{\epsilon}{||g_t(f(x_{t}^i))||_{p}})$.} Similar to the universal perturbation, we consider per-instance perturbations for both targeted and non-targeted attacks. The loss functions take the same form as the universal case defined in Eqs. \eqref{eq: non-targeted loss}-\eqref{eq: total targeted}.
	
	It is worth noting that MTA requires a lower storage cost when compared with previous generative approaches to craft adversarial examples. This is because MTA uses a shared encoder for all tasks, while previous generative approaches such as GAP \cite{DBLP:conf/cvpr/PoursaeedKGB18} needs a different encoder for each task. The usage of a shared encoder in MTA leads to faster inference when several tasks share the same inputs (\ie $a_1 = a_2 = a_3 = a$ in Figure \ref{fig:architechture}), since the encoded representation $f(a)$ only needs to be computed once and will be decoded by all tasks, while single-task generative methods compute $M$ different encoded representations independently. Moreover, MTA is faster than optimization-based attack methods at the inference phase, since the latter needs to solve an optimization problem for each test instance.

	\section{Experiments}
	In this section, we empirically evaluate the performance of the proposed MTA method.
	
	We conduct experiments on two datasets, including the {Office-31} dataset \cite{saenko2010adapting} and the NYUv2 dataset \cite{silberman2012indoor}. To the best of our knowledge, all the adversarial attack methods work under the single-task learning setting and here we choose the GAP method \cite{DBLP:conf/cvpr/PoursaeedKGB18} as a baseline since it is very relevant to the proposed MTA as discussed in Section \ref{sec:related_work_attack}. 
	
	In the experiments, we adopt the most widely used $L_{\infty}$ norm for the perturbations, \ie $p=\infty$. The uniform weights for tasks are used, \ie $\alpha_{t} = \frac{1}{M}$ in Eqs. \eqref{eq: total non-targeted} and \eqref{eq: total targeted}. In principle, adaptive weighting strategies \cite{sener2018multi,DBLP:conf/cvpr/KendallGC18} to learn $\{\alpha_t\}_{t=1}^M$ can be used to improve performance of MTA and it is left for the future work. All models are implemented via PyTorch \cite{paszke2017automatic} and trained with the Adam optimizer \cite{kingma2014adam}.
	For the experiments on the Office-31 dataset, we set the learning rate as $2e-4$ and training batch size as 10 for each task. The number of ResNet blocks in the generator is 6. For the experiments on the NYUv2 dataset, the same optimizer and learning rate are used. The training batch size is 5 for each task. We use 4 ResNet blocks for universal perturbations and 10 for per-instance perturbations, since we empirically found that generators in the the latter setting are more difficult to train and require more blocks.

	

	\subsection{Experiments on Office-31 Dataset} \label{sec: results office31}
	
	The {Office-31} dataset \cite{saenko2010adapting} consists of 4,110 images in 31 categories shared by three tasks: \textit{Amazon} (\textbf{A}) that contains images downloaded from amazon.com, \textit{Webcam} (\textbf{W}) and \textit{DSLR} (\textbf{D}) which are images taken by the web camera and digital SLR camera under different environmental settings.
	
	Three pre-trained classifiers are trained on tasks \textbf{A}, \textbf{D} and \textbf{W} independently, with clean accuracies $78.72\%, 98.30\%$ and $91.49\%$ respectively. The proposed MTA is to generate adversarial perturbations for all three pre-trained classifiers simultaneously. In the case of non-targeted attacks, high-quality perturbations should achieve a high fooling ratio and result in a low accuracy on adversarial examples. Given the pre-trained model $\mathcal{K}_t$, the fooling ratio is defined as the proportion of inputs $\mathbf{x}_t^{i}$ for which after the perturbation, $\mathcal{K}_t(\hat{\mathbf{x}}_t^{i})\neq \mathcal{K}_t(\mathbf{x}_t^{i})$ holds. In the case of targeted attacks, high-quality perturbations should result in a higher top-1 target accuracy, which is the proportion of the adversarial examples that are classified as the target label. 
	
	\subsubsection{Universal Perturbations}
	
	\paragraph{Non-targeted Universal Perturbations.} In this setting, we seek to find a universal perturbation $v_{t}$ for each task $t$ to decrease the overall performance of the pre-trained model $\mathcal{K}_{t}$. The results are shown in Table \ref{office31_universal_}. Note that when the norm of allowed perturbations is relatively large ($\epsilon = 10$), MTA and GAP have comparable performance. However, if we set $\epsilon = 5$, we see that MTA outperforms GAP on all tasks with respect to both the fooling ratios and accuracies of pre-trained models. This is probably because jointly learning multiple tasks provides shared knowledge that is useful for the training process, especially for smaller $\epsilon$ which corresponds to a more difficult attack. See Figure \ref{app_fig:office31_universal_unt} in Appendix for visualization of the generated perturbations.
	
	\begin{table}[!htbp]
		\centering
		\resizebox{\linewidth}{!}{
			\begin{tabular}{|c|c|c|c|c|c|c|}
				\hline
				\multirow{2}{*}{} & \multirow{2}{*}{} & \multicolumn{3}{c|}{Task} &\multirow{2}{*}{Avg}\\
				\cline{3-5}
				& &A&D&W&\\
				\hline
				\multirow{4}{*}{$\epsilon = 10$} &
				\multirow{2}{*}{GAP} & 97.83\% & \textbf{96.52\%} & \textbf{96.52\%} & \textbf{96.96\%} \\
				& & (2.61\%) & (\textbf{3.48\%}) & (3.48\%) & (3.19\%) \\
				\cline{2-6}
				& \multirow{2}{*}{MTA} & \textbf{98.26\%} & 96.09\% & \textbf{96.52\%} & 96.95\% \\
				& & (\textbf{2.17\%}) & (3.91\%) & (\textbf{2.61\%}) & (\textbf{2.90\%}) \\
				\cline{1-6}
				\multirow{4}{*}{$\epsilon = 5$} &
				\multirow{2}{*}{GAP} & 88.26\% & 75.65\% & 78.26\% & 80.72\% \\
				& & (12.61\%) & (24.35\%) & (20.87\%) & (19.28\%) \\
				\cline{2-6}
				& \multirow{2}{*}{MTA} & \textbf{90.21\%} &  \textbf{91.49\%} & \textbf{82.13\%} & \textbf{88.80\%} \\
				& & (\textbf{9.36\%}) & (\textbf{8.51\%}) & (\textbf{17.02\%}) & (\textbf{11.63\%}) \\
				\hline
		\end{tabular}}
		\vskip 0.1in
		\caption{The fooling ratios and the accuracies of pre-trained models (indicated in the parenthesis) for non-targeted universal perturbations on the Office-31 dataset. Better attack performance results (\ie higher fooling ratios and lower accuracies) are shown in \textbf{bold}.}
		\label{office31_universal_}
	\end{table}
	
	\paragraph{Targeted Universal Perturbations.} In this setting we would like to find a universal perturbation $v_{t}$ for each task $t$ such that most perturbed images will be classified as the target class desired by users. Table \ref{office31_universal_targeted} shows the top-1 accuracies for both $\epsilon = 5$ and $\epsilon = 10$ when the ``bike helmet" class is chosen as the target class for all three tasks. MTA outperforms GAP in all the cases except one (\ie task \textbf{D} when $\epsilon = 5$). We also experiment on choosing different target classes for different tasks and obtain consistent results (Table \ref{app_tbl:office31_universal_targeted} in the Appendix). 
	See Figure \ref{app_fig:office31_universal_targeted} in Appendix for the visualization of the perturbations.
	
	\begin{table}[!htbp]
		\centering
		\resizebox{\linewidth}{!}{
			\begin{tabular}{|c|c|c|c|c|c|c|}
				\hline
				\multirow{2}{*} & \multirow{2}{*} & \multicolumn{3}{c|}{Task} &\multirow{2}*{Avg}\\
				\cline{3-5}
				& &A&D&W&\\
				\hline
				\multirow{2}*{$\epsilon = 10$} &
				GAP & 96.52\% & 98.70\% & 98.70\% & 97.97\% \\
				& MTA & \textbf{99.15\%} & \textbf{100.00\%} & \textbf{99.57\%} & \textbf{99.57\%} \\
				
				\cline{1-6}
				\multirow{2}*{$\epsilon = 5$} &
				GAP & 76.09\% & \textbf{68.26\%} & 64.78\% & 69.71\% \\
				& MTA & \textbf{81.74\%} &  64.35\% & \textbf{65.22\%} & \textbf{70.44\%} \\
				\hline
		\end{tabular}}
		\vskip 0.1in
		\caption{The top-1 target accuracies for targeted universal perturbations on the Office-31 dataset. Higher top-1 targeted accuracies are shown in \textbf{bold}.}
		\label{office31_universal_targeted}
	\end{table}

	\subsubsection{Per-instance Perturbations}
	\paragraph{Non-targeted per-instance Perturbations.} In this setting, we train a generator which maps a certain input to its additive perturbation so that the overall performance of pre-trained models decreases. The fooling ratios as well as the accuracies of pre-trained models after perturbations are shown in Table \ref{office31_per-instance_}, showing that MTA performs better than GAP on average with both thresholds. When $\epsilon = 5$, we can see that MTA leads GAP by $15\%$ in task \textbf{D} in terms of the fooling ratios, while it performs worse than GAP in task \textbf{W}. One possible reason is that the inclusion of task \textbf{W} improves the performance of the other tasks in MTA, though the performance of task \textbf{W} is relatively poor. Moreover, we observe that the overall performance of non-targeted per-instance perturbations in Table \ref{office31_per-instance_} is not as good as that of non-targeted universal perturbations in Table \ref{office31_universal_}, especially when $\epsilon$ is small. Similar phenomenon can also be observed in \cite{DBLP:conf/cvpr/PoursaeedKGB18} on the ImageNet dataset under the single-task setting. This is probably because universal perturbations are easier to learn than per-instance ones by the generator in the classification tasks.
	
	\begin{table}[!htbp]
		\centering
		\resizebox{\linewidth}{!}{
			\begin{tabular}{|c|c|c|c|c|c|c|}
				\hline
				\multirow{2}{*} & \multirow{2}{*} & \multicolumn{3}{c|}{Task} &\multirow{2}*{Avg}\\
				\cline{3-5}
				& &A&D&W&\\
				\hline
				\multirow{4}*{$\epsilon = 10$} &
				\multirow{2}*{GAP} & 91.91\% & \textbf{97.87\%} & 96.17\% & 95.32\% \\
				& & (6.81\%) & (\textbf{2.13\%}) & (4.26\%) & (4.40\%) \\
				\cline{2-6}
				& \multirow{2}*{MTA} & \textbf{93.91\%} & 97.39\% & \textbf{96.52\%} & \textbf{95.94\%} \\
				& & (\textbf{4.78\%}) & (3.04\%) & (\textbf{2.61\%}) & (\textbf{3.48\%}) \\
				\cline{1-6}
				
				\multirow{4}*{$\epsilon = 5$} &
				\multirow{2}*{GAP} & 82.13\% & 65.53\% & \textbf{90.72\%} & 79.46\% \\
				& & (16.17\%) & (34.47\%) & (\textbf{9.28\%}) & (19.97\%) \\
				\cline{2-6}
				& \multirow{2}*{MTA} & \textbf{83.04\%} &  \textbf{80.43\%} & 81.30\% & \textbf{81.59\%} \\
				& & (\textbf{16.09\%}) & (\textbf{19.57\%}) & (18.26\%) & (\textbf{17.97\%}) \\
				\hline
			\end{tabular}
		}
		\vskip 0.1in
		\caption{The fooling ratios and accuracies of pre-trained models (indicated in the parenthesis) for non-targeted per-instance perturbations on the Office-31 dataset. Better attack performance results (\ie higher fooling ratios and lower accuracies) are shown in \textbf{bold}.}
		
		\label{office31_per-instance_}
	\end{table}

\begin{figure}[!htbp]
		\centering
		\includegraphics[width=1\linewidth]{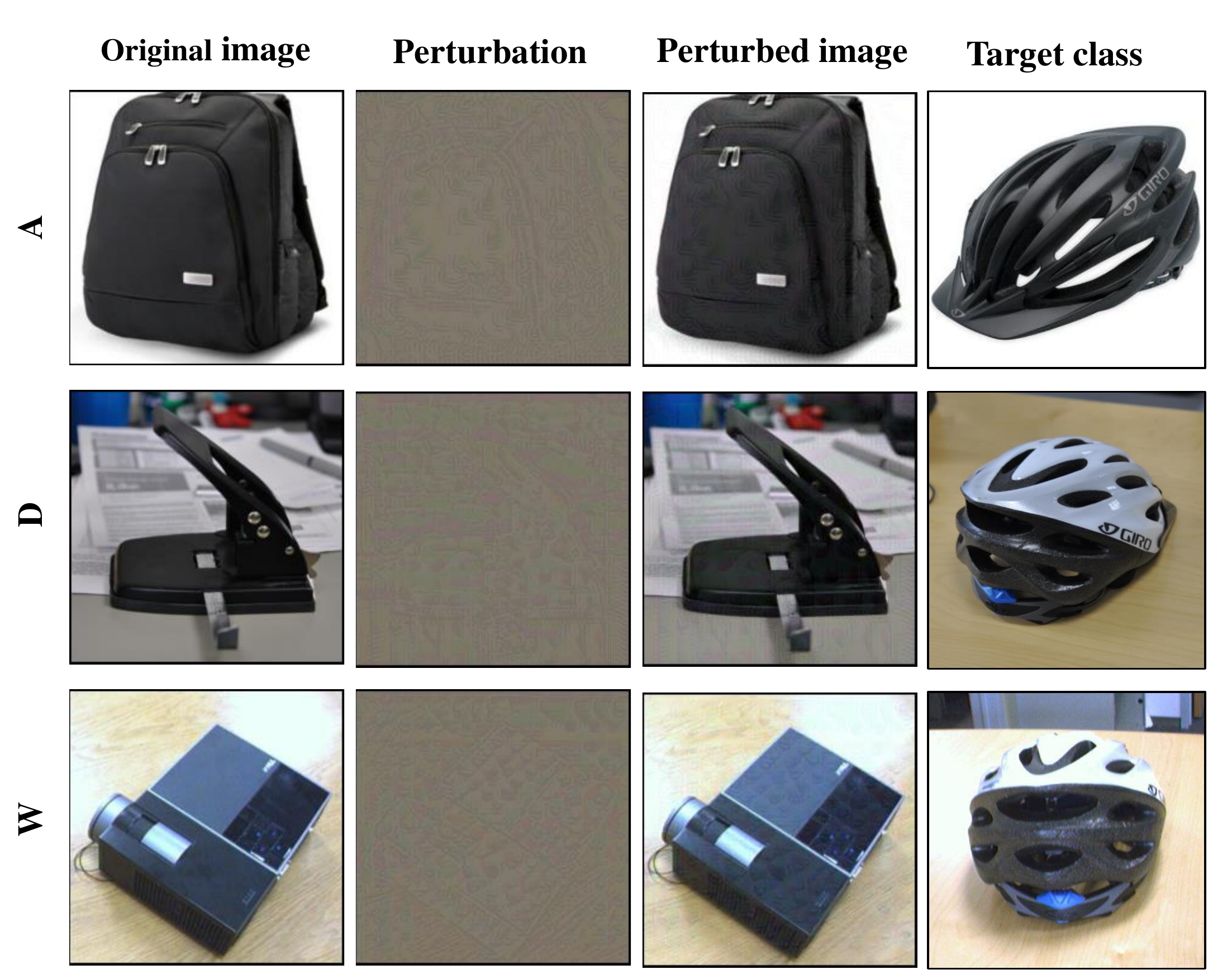}
		\caption{Targeted per-instance perturbations generated with $\epsilon = 5$ on the Office-31 dataset.}
		\label{fig:office31_imdep_targeted}
	\end{figure}

	\paragraph{Targeted Per-instance Perturbations.}
	In this setting, we train a generator which maps an input to its additive noise such that the resulting adversarial example is classified as the target class (\ie ``bike helmet") specified by the attackers. According to the results of the top-1 accuracies shown in Table \ref{office31_per-instance_targeted}, we find that MTA outperforms GAP on average and for almost all the tasks, which verifies the effectiveness of the proposed MTA. In Figure \ref{fig:office31_imdep_targeted}, generated perturbations are visualized. By closely inspecting the perturbations, we can see that the generated perturbations have similar shapes to the corresponding input images as well as similar texture to the target class.
	\begin{table}[!htbp]
		\centering
		\resizebox{\linewidth}{!}{
			\begin{tabular}{|c|c|c|c|c|c|c|}
				\hline
				\multirow{2}{*} & \multirow{2}{*} & \multicolumn{3}{c|}{Task} &\multirow{2}*{Avg}\\
				\cline{3-5}
				& &A&D&W&\\
				\hline
				\multirow{2}*{$\epsilon = 10$} &
				GAP & 85.96\% & 91.06\% & 94.89\% & 90.64\% \\
				& MTA & \textbf{88.70\%} & \textbf{95.22\%} & \textbf{95.65\%} & \textbf{93.19\%} \\
				
				\cline{1-6}
				\multirow{2}*{$\epsilon = 5$} &
				GAP & \textbf{60.43\%} & 74.68\% & 45.99\% & 60.37\% \\
				& MTA & 54.04\% &  \textbf{77.45\%} & \textbf{51.49\%} & \textbf{60.99\%} \\
				\hline
		\end{tabular}}
		\vskip 0.1in
		\caption{Top-1 target accuracies for targeted per-instance perturbations on the Office-31 dataset. Higher top-1 targeted accuracies is shown in \textbf{bold}.}
		\label{office31_per-instance_targeted}
	\end{table}
	
	\begin{table*}[!htb]
		\centering
		\begin{tabular}{cccccccccccc}
			\toprule
			\multirow{3}{*}{Type} & \multirow{3}{*}{Attack} & \multicolumn{2}{c}{Segmentation} & \multicolumn{2}{c}{Depth} & \multicolumn{5}{c}{Surface Normal}  \\
			\cmidrule(r){3-4} \cmidrule(r){5-6} \cmidrule(r){7-11}
			& & \multirow{2}{*}{mIOU $\uparrow$} &  \multirow{2}{*}{Pix Acc $\uparrow$} &  \multirow{2}{*}{Abs Err $\downarrow$} &  \multirow{2}{*}{Rel Err$\downarrow$} & \multicolumn{2}{c}{Angle Distance $\downarrow$} & \multicolumn{3}{c}{Within $t$\degree $\uparrow$} \\ \cmidrule(r){7-8} \cmidrule(r){9-11} & & & & & &  Mean  & Median   & 11.25  & 22.5  & 30   \\
			\midrule
			
			Clean & &  18.63 &  53.50  & 0.6298  & 0.2500  &  33.01 &  28.91  & 17.77 & 39.26 & 51.92 \\
			\cmidrule(r){1-11}
			\multirow{2}{*}{Universal} &  GAP  & 5.19 & 19.86 & 1.6410 & 0.5518 &  \textbf{69.45} & \textbf{72.62} & 4.09 & 10.52 & 15.02 \\
			& MTA  & \textbf{5.09} & \textbf{19.23} & \textbf{1.6507} & \textbf{0.5565} & 68.61 & 71.57 & \textbf{3.66} & \textbf{10.00} & \textbf{14.62} \\
			\cmidrule(r){1-11}
			\multirow{2}{*}{Per-instance} & GAP & 1.34 &  6.29 & \textbf{1.8507} & \textbf{0.6263} & 69.23 & 72.49 & 2.26 & 7.19 & 11.27 \\
			& MTA & \textbf{1.29} & \textbf{6.27} & 1.7962 & 0.6045 & \textbf{69.71} & \textbf{73.43} & \textbf{1.97} & \textbf{6.67} & \textbf{10.76} \\
			\bottomrule
		\end{tabular}
		\vskip 0.1in
		\caption{The performance of pre-trained model under attacks with $\epsilon = 5$ on the NYUv2 dataset. $\uparrow$ ($\downarrow$) means the higher (lower), the better the metrics and the worse quality of the perturbations. ``Clean" means clean data, ``Universal" means universal attacks and ``Per-instance" means per-instance attacks. Better attack performance results are shown in \textbf{bold}.}
		\label{tab:nyuv2_attack}
	\end{table*}

	\subsection{Experiments on NYUv2 Dataset} \label{sec: NYU results}

	The NYUv2 dataset \cite{silberman2012indoor} consists of RGB-D indoor scene images, each of which is used for three tasks: 13-class semantic segmentation \cite{couprie2013indoor}, depth estimation, and surface normal estimation \cite{eigen2015predicting}. By following \cite{liu2019end}, all training and validation images are resized to $288\times 384$ resolution to speed up training. We generate adversarial examples for fooling a Deep Multi-Task Learning (DMTL) network \cite{caruana97,zllt14,lmzcl15,zlzskyj15,mstgsvwy15,llc15} pre-trained on the NYUv2 dataset, which shares the first several layers as the shared encoder for all the tasks. 
	Since all tasks in this dataset are not standard classification tasks, we only perform non-targeted attacks and revise the loss function in Eq. \eqref{eq: non-targeted loss} as
	
	\begin{equation*} \label{eq: new non-targeted loss}
	\ell^{\mathrm{non-tar}}_{t} = \frac{1}{n_t}\sum_{i=1}^{n_t}-\log(\mathcal{L}_{t}(k_{t}(\hat{x}_{t}^i), {y_t^i})),
	\end{equation*}
	where $y_{t}^{i}$ is the ground truth and $\mathcal{L}_{t}$ is the pixel-wise cross-entropy loss, the $L_{1}$ loss and the element-wise dot product loss for semantic segmentation, depth estimation and surface normal estimation respectively. See \cite{liu2019end} for more details on these loss functions.

	\begin{figure*}[!htbp]
		\centering
		\includegraphics[width=1\linewidth]{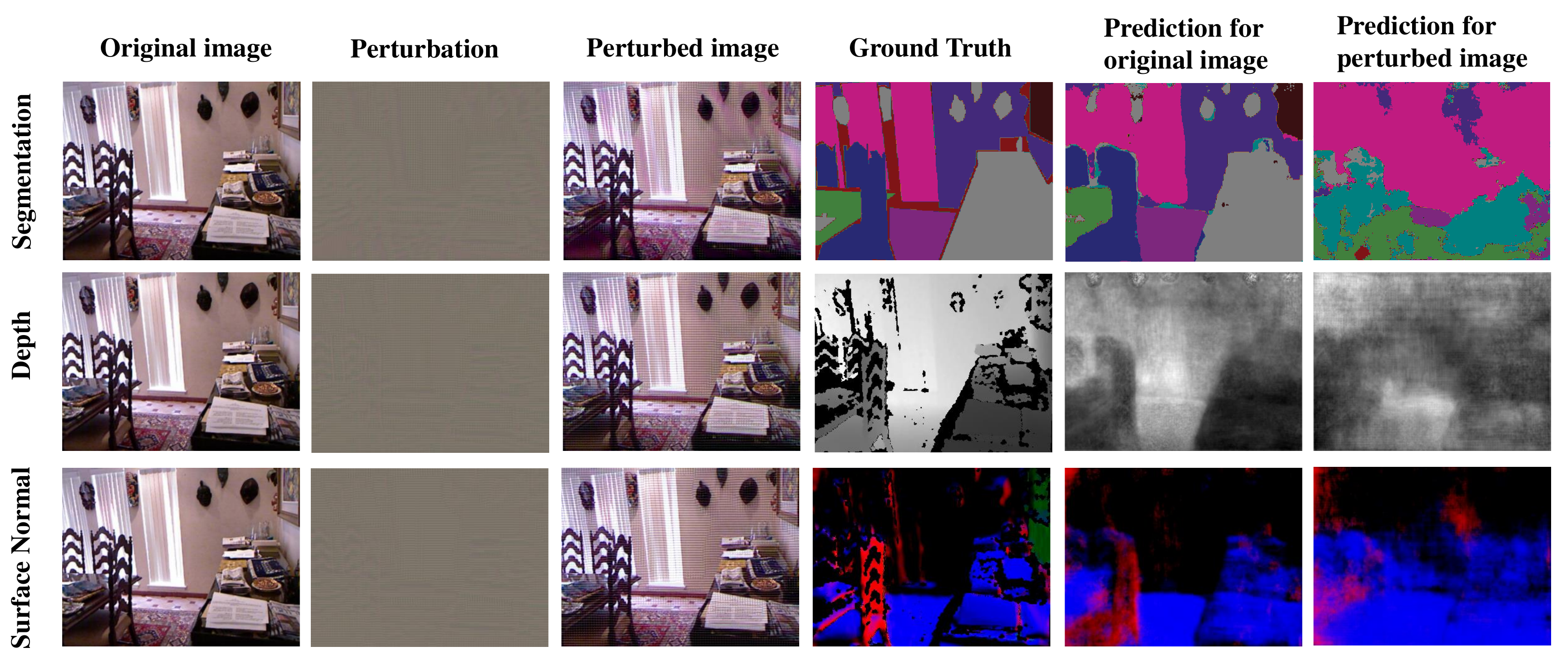}
		\caption{Per-instance perturbations with $\epsilon = 5$ on the NYUv2 dataset.}
		\label{fig:nyu_imdep}
	\end{figure*}

	\paragraph{Universal Perturbations.}
	In this setting, MTA learns $v_{t}$ for all tasks $t$ jointly by training a generator that exploits the shared knowledge among tasks. 
	Results shown in Table \ref{tab:nyuv2_attack} demonstrate that the proposed MTA method outperforms the single-task counterpart in most cases. We also visualize the perturbations and the resulting predictions in Figure \ref{app_fig:nyu_universal} in Appendix, showing that the generated perturbations successfully mislead the pre-trained model.
	

	\begin{table*}[!htb]
		\centering
		\begin{tabular}{cccccccccc}
			\toprule
			\multirow{3}{*}{Type} & \multicolumn{2}{c}{Segmentation} & \multicolumn{2}{c}{Depth} & \multicolumn{5}{c}{Surface Normal}  \\
			\cmidrule(r){2-3} \cmidrule(r){4-5} \cmidrule(r){6-10}
			& \multirow{2}{*}{mIOU $\uparrow$} &  \multirow{2}{*}{Pix Acc $\uparrow$} &  \multirow{2}{*}{Abs Err $\downarrow$} &  \multirow{2}{*}{Rel Err$\downarrow$} & \multicolumn{2}{c}{Angle Distance $\downarrow$} & \multicolumn{3}{c}{Within $t$\degree $\uparrow$} \\ \cmidrule(r){6-7} \cmidrule(r){8-10} & & & & & Mean  & Median   & 11.25  & 22.5  & 30   \\
			\midrule
			
			Clean &  21.00 &  55.88  & 0.6370 &  0.2632 &   30.62 &  26.19 &  20.48 & 43.68 & 56.54 \\
			\cmidrule(r){1-10}
			Universal  & 9.89 & 31.67 &  1.3244 & 0.4418  &  \textbf{47.83} & \textbf{45.84} &  7.33 &  21.00 & 30.57 \\
			Per-instance  & \textbf{8.84} &  \textbf{28.48} & \textbf{1.4135} & \textbf{0.4689}  &  46.65 & 45.00 &  \textbf{6.01} &  \textbf{19.19} & \textbf{29.35} \\
			\bottomrule
		\end{tabular}
		\vskip 0.1in
		\caption{Transferability of universal and per-instance perturbations on the NYUv2 dataset with $\epsilon = 5$. The multi-task perturbation generator is trained to fool the DMTL model and is tested on the MTAN model. $\uparrow$ ($\downarrow$) means the higher (lower), the better the metric and the worse quality of perturbations. Better attack performance results are shown in \textbf{bold}.}
		\label{tab:nyuv2_transfer}
	\end{table*}

	\paragraph{Per-instance Perturbations.} In this setting, we aim to find a generator which maps an input image to its corresponding perturbation. The results are given in Table \ref{tab:nyuv2_attack}, showing that MTA outperforms GAP in most cases. Different from previous experiments on the Office-31 dataset, here we observe that per-instance perturbations outperform universal perturbations in all tasks on the NYUv2 dataset. This is probably because unlike the Office-31 dataset, tasks in the NYUv2 dataset share the same inputs, which makes it easier to learn per-instance perturbations. Figure \ref{app_fig: attack_strength_change} in the Appendix further validates this , where more perturbation thresholds are considered. It also shows that larger perturbation thresholds result in stronger attacks. Finally, we visualize per-instance perturbations in Figure \ref{fig:nyu_imdep}. By closely inspecting the perturbations, we can observe that shapes of perturbations resemble the original images. Also, from Figure \ref{fig:nyu_imdep}, we can see that the performance of the pre-trained model significantly drops on the adversarial examples.

	\subsection{Comparison on Inference Time and Storage}
	
	As discussed in Section \ref{sec: per-instance pert} , MTA can reduce the inference time and storage cost compared with its single-task counterpart. Table \ref{nyu_par_time} gives comparison of MTA and GAP in terms of the number of parameters and inference time. From the results, we can see that MTA has a lower storage cost and makes inference faster than GAP, which is due to the shared encoder in MTA for multiple tasks.


	\begin{table}[!htbp]
		\centering
		\scalebox{1}{
			\begin{tabular}{cccc}
				\hline
				& & P(M) & T(ms) \\
				\hline
				\multirow{2}*{Universal} &
				GAP & 1.64 & 384 \\
				& MTA & \textbf{0.62} & \textbf{137}  \\
				\cline{1-4}
				\multirow{2}*{Per-instance} &
				GAP & 3.77 & 342 \\
				& MTA & \textbf{1.33} &  \textbf{126}  \\
				\hline
		\end{tabular}}
		\vskip 0.1in
		\caption{Comparison of MTA and GAP in terms of number of parameters (P) and inference time (T) for three tasks on the NYUv2 dataset. The inference time refers to the time for generating three perturbations, one for each task.}
		\label{nyu_par_time}
	\end{table}
	
	\subsection{Transferability}
	
	Previous works on single-task adversarial attacks have demonstrated that adversarial examples generated against one model can often mislead other deep learning models trained on the same dataset, and this property is referred to as the transferability \cite{DBLP:journals/corr/SzegedyZSBEGF13, DBLP:conf/nips/IlyasSTETM19}. The transferability can be leveraged to achieve black-box attacks \cite{DBLP:journals/corr/PapernotMGJCS16,DBLP:journals/corr/GoodfellowSS14}, where the attacker needs not to have the knowledge of the pre-trained models, including parameters and architectures. In this section, we test the transferability of the perturbations generated by MTA against DMTL to another deep multi-task learning model, MTAN \cite{liu2019end}, both of which are pre-trained on the NYUv2 dataset, and show the results in Table \ref{tab:nyuv2_transfer}. According to the results, we can see that the perturbations against DMTL also greatly decrease the performance of MTAN and this verifies that the proposed MTA can generate adversarial examples with good transferability.

	\section{Conclusion and Future Work}
	In this paper, we extend adversarial attacks to the multi-task setting by proposing the MTA that, when learning to generate perturbations, exploits the shared knowledge among tasks by parameter sharing. We demonstrate that the proposed MTA method can generate perturbations of higher-quality and reduce inference time as well as storage cost. In future work, we are interested in designing defence models for the MTA.
	
\bibliographystyle{ieee_fullname}
\bibliography{MTA}

\end{document}